%% file: 00_main.tex
\documentclass[letterpaper, 10 pt, conference]{ieeeconf}  

\IEEEoverridecommandlockouts                              

\input{99_macros}

\title{\LARGE \bf
\methodname: End-to-end Gaze Target Detection with Head-Target Association}

\author{Zhi-Yi Lin$^{1}$, Jouh Yeong Chew$^{2}$, Jan van Gemert$^{1}$, Xucong Zhang$^{1}$
\thanks{$^{1}$Computer Vision Lab, Delft University of Technology}%
\thanks{$^{2}$Honda Research Institute Japan}%
}

\begin{document}

\maketitle

\input{01_abstract}
\input{02_introduction}
\input{03_relatedwork}
\input{04_method}
\input{05_experiment}

\input{07_conclusion}


%
%
\bibliographystyle{IEEEtran}
\bibliography{IEEEabrv, 08_reference}

\end{document}

%% file: 99_macros.tex
\usepackage{graphicx}
\usepackage{booktabs}

\usepackage[accsupp]{axessibility}  

\usepackage{amsmath}
\usepackage{amssymb}
\usepackage{multirow}
\usepackage{subcaption}
\usepackage{tabularx}
\usepackage{tabulary}
\usepackage{adjustbox}
\usepackage{xcolor}

\usepackage{graphicx}
\usepackage{pifont}
\usepackage{bbm}
\usepackage{dsfont}
\usepackage{makecell}
\usepackage{xspace}
\usepackage{cite}

\usepackage[pagebackref,breaklinks,colorlinks]{hyperref}

\usepackage{orcidlink}

\usepackage[capitalize]{cleveref}
\crefname{section}{Sec.}{Secs.}
\Crefname{section}{Section}{Sections}
\Crefname{table}{Table}{Tables}
\crefname{table}{Tab.}{Tabs.}

\newcommand{\methodname}{GazeHTA\xspace}

\newcolumntype{P}[1]{>{\raggedright\arraybackslash}m{#1}}%
\newcolumntype{C}[1]{>{\centering\arraybackslash}m{#1}}%
\newcolumntype{R}[1]{>{\raggedleft\arraybackslash}m{#1}}%
\newcolumntype{Y}{>{\centering\arraybackslash}X}

\definecolor{xucongcolor}{rgb}{0.73725, 0.6588, 0.0705} 
\newif\ifshowcomments
\showcommentstrue 

\usepackage[normalem]{ulem}
\newif\ifshowchange

\def\etal{et~al.\xspace}

\usepackage{textcomp}
\usepackage{gensymb}
\usepackage{diagbox}

\newcommand{\cmark}{\ding{51}}%
\newcommand{\xmark}{\ding{55}}%

%% file: 01_abstract.tex
\begin{abstract}
Precisely detecting which object a person is paying attention to is critical for human-robot interaction since it provides important cues for the next action from the human user.
We propose an end-to-end approach for gaze target detection: predicting a head-target connection between individuals and the target image regions they are looking at. Most of the existing methods use independent components such as off-the-shelf head detectors or have problems in establishing associations between heads and gaze targets. In contrast, we investigate an end-to-end multi-person \textbf{Gaze} target detection framework with \textbf{H}eads and \textbf{T}argets \textbf{A}ssociation (\textit{\methodname}), which predicts multiple head-target instances based solely on input scene image. \methodname addresses challenges in gaze target detection by (1) leveraging a pre-trained diffusion model to extract scene features for rich semantic understanding, (2) re-injecting a head feature to enhance the head priors for improved head understanding, and (3) learning a connection map as the explicit visual associations between heads and gaze targets. Our extensive experimental results demonstrate that \methodname outperforms state-of-the-art gaze target detection methods and two adapted diffusion-based baselines on two standard datasets. 
\end{abstract}

%% file: 02_introduction.tex
\section{Introduction}
\label{sec:intro}

A collaborative robot must quickly and accurately understand the human user intentions \cite{admoni2017social,schellen2021robot,sheikhi2015combining}. Among non-verbal cues, human eye gaze is a critical indicator of both intention \cite{arreghini2024predicting} and attention \cite{oishi20214d}. Eye gaze has been extensively applied in human-robot interaction for tasks such as hand motion prediction \cite{jayasuriya20243d}, robot navigation \cite{zhang2023human,holman2021watch}, pick-and-place actions \cite{prada2023gaze,razali2022using}, object referencing \cite{qian2023gvgnet}, walking assistance \cite{zhang2021foot}, robotic manipulation \cite{kim2020using}, and facilitating the transfer of objects between humans and robots \cite{kshirsagar2020robot}.
In gaze estimation, a common approach is to infer the 3D gaze direction from the eyes \cite{zhang2017mpiigaze,yu2018deep,zhou2017two}, face \cite{zhang2017mpiigaze,cheng2023dvgaze,bandi2023new}, or full-body joints \cite{kratzer2020mogaze}. However, many real-world applications demand not just the gaze direction, but the precise identification of the region or object a person is looking at—referred to as the gaze target.

In contrast to conventional gaze estimation methods, Gaze target detection aims to directly associate individuals and their gaze targets within a single image \cite{recasens2015they} or across multiple video frames \cite{recasens2017following}, offering an end-to-end solution for human attention estimation.
Most of the gaze target detection approaches employ two-stream architectures, where one stream focuses on scene feature extraction while the other one learns head features \cite{chong2018connecting,lian2018believe, chong2020detecting,bao2022escnet}. These approaches face challenges such as the lack of direct association between heads and gaze targets, and the dependency on off-the-shelf head detectors as demonstrated in recent studies \cite{tu2022end,tonini2023object}.
Moreover, the majority of previous methods are restricted to processing one head at a time, requiring repeated processing to identify gaze targets for all individuals in the scene when multiple people are present.
Tu \etal~\cite{tu2022end} addresses the challenges in the two-stream approaches with an end-to-end multi-person gaze target detection method, to directly predict multiple pairs of head-target instances. Nevertheless, this approach does not establish strong connections between heads and gaze targets, which restricts the final detection accuracy. Tonini \etal~\cite{tonini2023object} improves the association between heads and gaze targets, and achieves state-of-the-art performance. It is, however, not fully end-to-end due to its dependency on preliminary object detection. Pre-training the object detector also limits the object categories considered for gaze targets. This constraint may exclude diverse object classes or regions within the environment that are beyond the scope of the pre-trained object detector.

\begin{figure}[t]
  \centering
  \includegraphics[width=\linewidth]{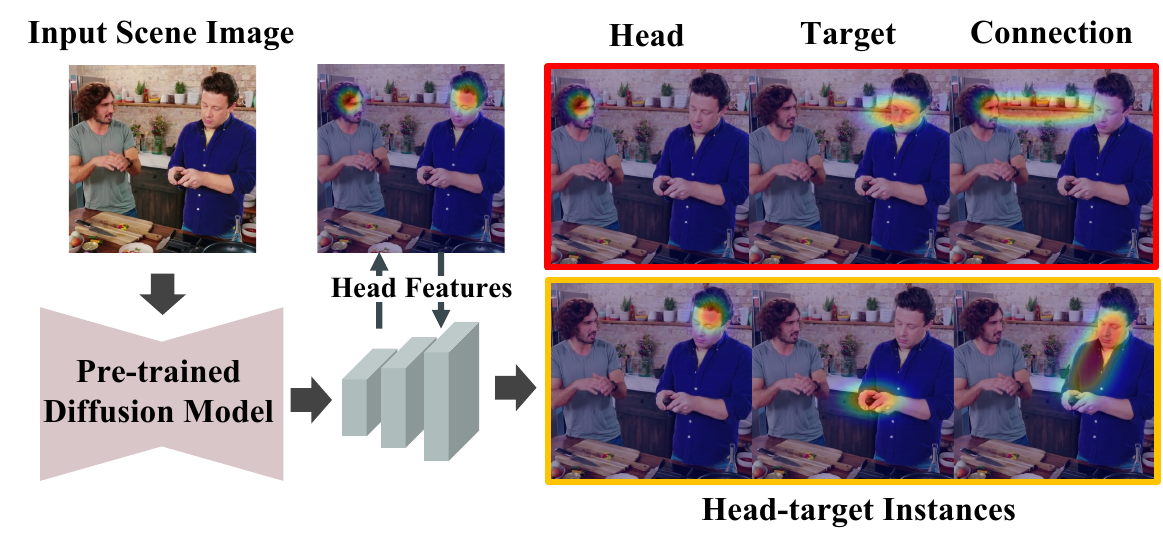}
  \caption{\methodname takes the scene image as input to predict head-target instances. \methodname consists of a pre-trained diffusion model as the scene feature extractor, a head feature re-injection mechanism, and the connection maps as the visual associations between heads and gaze targets.
  }
  \label{fig:overview}
\end{figure}

There are two major challenges in the gaze target detection task: 1) detecting the heads and the gaze targets; and 2) associating heads and gaze targets to predict which head is looking at which target. Addressing the first challenge involves identifying the relevant heads, objects, or other regions as potential gaze targets. This can be solved by using powerful feature extractors, such as diffusion-based models, which have demonstrated remarkable performance in various visual perception tasks~\cite{chen2023diffusiondet,ji2023ddp,rombach2022high,saxena2024surprising,zhao2023unleashing}. However, effectively integrating these pre-trained large foundation models into gaze target detection remains an open question.
The second challenge entails linking heads and corresponding gaze targets.
Despite efforts to employ gaze estimation results as constraints on the gaze target locations \cite{lian2018believe,fang2021dual,tonini2023object,tafasca2023childplay}, most existing methods have not fully addressed the association between heads and gaze targets.

In this paper, we propose \methodname: a novel end-to-end framework for multi-person gaze target detection. 
We exploit features from a pre-trained diffusion model, while emphasizing head regions with a novel re-injection method. 
We then associate the heads and gaze targets explicitly with a connection map in the image. These connection maps serve as links between the heads and gaze targets in the image, offering additional supervision during model training.
An overview is shown in \cref{fig:overview}.
Our experiments demonstrate that \methodname achieves state-of-the-art performance while not being restricted to the diffusion model backbone.
In summary, the main contributions of this paper include:
\begin{itemize}
    \item We are the first to exploit the rich semantic features from the pre-trained diffusion model for the gaze target detection task.
    \item We propose a head feature re-injection to improve the head priors, and a connection map to explicitly associate heads and gaze targets.
    \item Our extensive experiments show that \methodname achieves state-of-the-art performances on standard datasets.
\end{itemize}

%% file: 03_relatedwork.tex
\section{Related Works}
\subsection{Gaze Target Detection}
Formulated by \cite{recasens2015they}, gaze target detection is to determine the location where a person is looking within a scene image, or across video frames~\cite{recasens2017following}.
A two-stream approach is a common strategy with one stream focusing on scene understanding, while another one learning head features from input head crops, detected by an off-the-shelf head detector \cite{chong2018connecting,lian2018believe,chong2020detecting,tafasca2023sharingan}. Various modalities such as temporal information \cite{chong2020detecting}, depth \cite{fang2021dual,bao2022escnet,tafasca2023childplay,tonini2022multimodal}, and 2D pose \cite{bao2022escnet} have been incorporated to increase accuracy. However, relying on off-the-shelf models for head crops or incorporating multiple modalities remains challenging, particularly in ensuring model robustness for real-world scenarios. Studies have shown a significant performance drop when the input head crops are detected by off-the-shelf detectors instead of human annotations \cite{tu2022end,tonini2023object}.

Recently, end-to-end approaches have emerged to simultaneously detect multiple heads and their associated gaze targets \cite{tu2022end,tonini2023object}.
The HGTTR is a transformer-based network that predicts several head-target instances comprising head locations, out-of-frame flags, and gaze heatmaps \cite{tu2022end}. Unfortunately, its performance suffers from brittle associations between heads and gaze targets.
As a result, the latest advancements 
leverage the concept of objects to establish associations between heads and gaze targets \cite{tonini2023object,wang2024transgop}. These object-aware frameworks initially predict objects and subsequently associate head objects with all detected objects.
However, these methods revert to a two-stage approach because of their reliance on detected objects for head-object association. 
Moreover, the scope of gaze targets is limited to the classes that the pre-trained object detector was trained.
To efficiently learn the correlation between heads and gaze targets, it has been shown that explicitly connecting the head and target with a heatmap is feasible \cite{zhao2020learning} while the potential of such an explicit connection has not been explored yet.
In contrast to previous approaches, \methodname offers an end-to-end solution with implicit target feature extraction, enhanced head priors, and finally, an explicit association between heads and gaze targets through connection maps. 

\subsection{Diffusion Models}
Originally from diffusion probabilistic models \cite{sohl2015deep}, diffusion models are designed to learn the distribution of data for generating new samples.
This process involves iteratively applying transformations from a noise image to a realistic image \cite{rombach2022high,ramesh2021dalle,saharia2022photorealistic, ramesh2021zero}.

With the remarkable success of diffusion models in image generation, recent studies showed their potential across various downstream visual perception tasks such as object classification~\cite{clark2024text} and depth estimation~\cite{saxena2024surprising}, which are strongly correlated with the gaze target detection task. 
It is straightforward to extend to gaze target detection that the scene image serves as the condition for predicting head and gaze heatmaps through multi-step denoising. However, the output of gaze target detection is heatmaps instead of high-quality images. Therefore, directly applying the denoising process may not be efficient for gaze target detection.

Despite its simplicity, employing a pre-trained diffusion model as a feature extractor is effective for various visual perception tasks, including image segmentation and depth estimation \cite{zhao2023unleashing,yang2024depth}.
Considering the output format of gaze target detection is a heatmap which resembles that of image segmentation and depth estimation, namely the spatial maps or heatmaps, they may potentially benefit from the same model architecture.
Following the same strategy, \methodname employs a pre-trained diffusion model as the backbone to extract both high-level and low-level semantic features for object understanding and head localization, which are crucial for accurate gaze target detection.

%% file: 04_method.tex
\section{\methodname}



\begin{figure*}[t]
  \centering
  \includegraphics[width=0.9\linewidth]{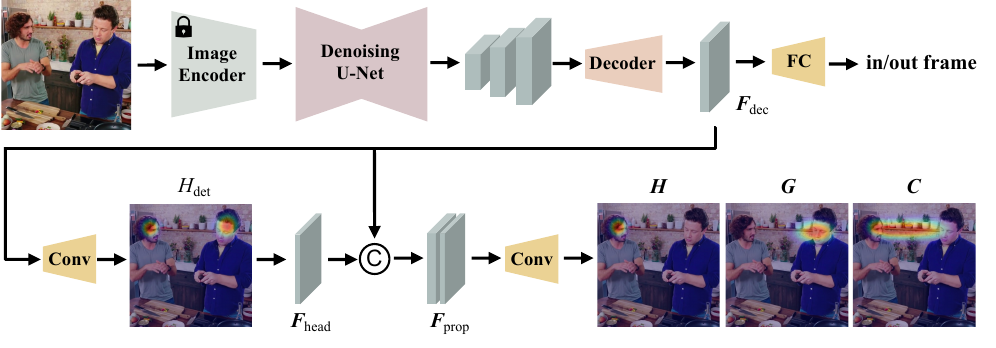}
  \caption{Network architecture of \methodname. The input scene image is transformed into scene features through a pre-trained image encoder and a denoising U-Net. The scene features are encoded for out-of-frame predictions via a fully connected layer, and for head-target proposals. In the head-target prediction branch, a learned head feature $\textbf{F}_\text{head}$ is re-injected and fused with the decoded feature $\textbf{F}_\text{dec}$. The resulting feature $\textbf{F}_\text{prop}$ is then used to predict $N$ head-target proposals, including head heatmaps $\textbf{H}$, gaze heatmaps $\textbf{G}$, and connection maps $\textbf{C}$, which are explicitly learned to associate the head and gaze target. Conv represents convolutional layers.}
  \label{fig:network}
\end{figure*}

The pipeline of \methodname is shown in \cref{fig:network}.
Taking only a scene image as the input, \methodname extracts multi-scale scene features for out-of-frame binary predictions and head-target proposals.
For the head-target proposal, a head feature is learned and re-injected back to enhance the head priors. Subsequently, heads and gaze targets are predicted in the form of heatmaps. Within each head-target instance, \methodname establishes a visual association between the head and its corresponding gaze target by a connection map.

\subsection{Scene Feature Extraction}
In \methodname, the backbone to extract scene features is a pre-trained denoising U-Net from Stable Diffusion \cite{rombach2022high}. The use of Stable Diffusion, a large-scale text-to-image generator, is due to its capability to encompass features with both low-level and high-level semantic knowledge \cite{pnvr2023ld} acquired from large-scale real-world data.
By taking the scene image as the only input, \methodname encodes it with a pre-trained image encoder, followed by a denoising U-Net from the pre-trained diffusion model.
We extract intermediate features from four different levels in the U-Net with a similar operation to the VPD \cite{zhao2023unleashing}.
These multi-scale scene features are then encoded into a compact representation 
for both out-of-frame predictions and head-target proposals. We expect the implicit understanding of the scene features can facilitate the identification of potential heads and gaze targets.
Note that, unlike VPD~\cite{zhao2023unleashing}, we eliminate the entire text prompt module by removing the cross-attention layers in the denoising U-Net, as it is not necessary nor effective for the gaze target detection task.

\subsection{Head Feature Re-injection}
Accurately locating the human head position is essential for successful gaze target detection. To improve head detection, we introduce supervision on a head detection map, which aims to detect all instances of heads within the scene image.
To this end, the multi-scale scene features 
are passed through a convolution-based decoder that gradually expands the spatial dimension 
into a decoded feature, denoted as $\textbf{F}_\text{dec} \in \mathbb{R}^{192 \times 64 \times 64}$. $\textbf{F}_\text{dec}$ is then passed through convolutional layers to generate an intermediate head detection map $H_\text{det}$ for head prior enhancement. To supervise the training on $H_\text{det}$, we apply the detection results from an off-the-shelf head detector \footnote{https://github.com/SibiAkkash/yolov5-crowdhuman} as pseudo labels since the heads are not fully annotated in the training data. Although $H_\text{det}$ provides insights into potential head positions, we only use it as a soft guidance for later head-target prediction.
To this end, we utilize the head priors encoded in its subsequent head feature $\textbf{F}_\text{head}$, which is derived by passing $H_\text{det}$ through a single convolutional layer.
After obtaining head priors, $\textbf{F}_\text{head}$ is re-injected and fused with $\textbf{F}_\text{dec}$ via concatenation to provide additional guidance for head localization. The resulting feature, denoted as $\textbf{F}_\text{prop} \in \mathbb{R}^{193 \times 64 \times 64}$ is utilized to predict $N$ head-target proposals.

\subsection{Visual Association via Connection Maps}
Given the scene feature for locating potential heads and targets, we use a visual association scheme via multi-task learning as the model not only predicts heads and targets but also connection maps as heatmaps.
Taking $\textbf{F}_\text{prop}$ as input, a prediction head with convolutional layers generates $N$ head-target proposals. Each proposal comprises three types of heatmaps: head heatmaps $\textbf{H} \in \mathbb{R}^{N \times 64 \times 64}$, the corresponding gaze heatmaps $\textbf{G} \in \mathbb{R}^{N \times 64 \times 64}$, and the connection maps $\textbf{C} \in \mathbb{R}^{N \times 64 \times 64}$. 
In contrast to previous approaches that typically link heads and gaze targets through a gaze vector prediction or the predefined field of view parameters \cite{lian2018believe,fang2021dual,tonini2023object, tafasca2023childplay}, the proposed connection maps $\textbf{C}$ explicitly learn the link between the heads and targets.
This is motivated by the fact that the prediction of head heatmaps $\textbf{H}$ and gaze heatmaps $\textbf{G}$ are isolated, so it is essential to have these connection maps $\textbf{C}$ to force the model to learn their associations \cite{zhao2020learning}.
It is important to note that the connection maps $\textbf{C}$ are a unidirectional mapping that only focuses on highlighting the connection between the head and its corresponding target. 

\subsection{Out-of-frame Flag}
Similar to previous gaze target detection works \cite{tu2022end,tonini2023object,tafasca2023childplay}, \methodname predicts an out-of-frame flag to indicate whether the person is looking outside the frame, in which case the predicted gaze targets are not considered. Same as the head-target proposals, the network utilizes $\textbf{F}_\text{dec}$ to predict $N$ out-of-frame flags, denoted as $O \in \mathbb{R}^{N}$, using a convolutional layer followed by a single fully-connected layer.
Although we can use the feature from a later stage, we empirically found that using $\textbf{F}_\text{dec}$ for the out-of-frame prediction yields the best overall performance in the final head-target instance prediction. 

\subsection{Training Objective}
Bipartite matching is first performed to find the optimal matching between the predicted head-target instances and the ground-truth instances. The loss is then computed between the matched instance pairs.

\subsubsection{Bipartite matching.}
Following the training procedure of \cite{carion2020end,tu2022end,tonini2023object}, we use the Hungarian algorithm \cite{kuhn1955hungarian} for the bipartite matching between the predicted head-target instances and the ground-truth. The bipartite matching finds the best one-to-one matching by minimizing the total matching cost between each ground-truth instance and the candidate predicted head-target instance. The computation of the matching cost for each potential pairing entails a weighted combination of the L2-norm of the gaze heatmaps $\textbf{G}$, the head heatmaps $\textbf{H}$, along with the L1-norm of the out-of-frame flags $O$. 
The weights for bipartite matching of $\textbf{G}$, $\textbf{H}$, and $O$ are empirically set to 1.0, 2.5, and 1.0, respectively.

\subsubsection{Loss function.}
The total loss is computed using the matched pairs of predicted and ground-truth instances, and is expressed as follows:
\begin{equation} 
\begin{aligned}
\mathcal{L}_{total} =  \lambda_{h} \cdot \mathcal{L}_{h} + \lambda_{g} \cdot \mathcal{L}_{g} + \lambda_{c} \cdot \mathcal{L}_{c} + \lambda_{o} \cdot \mathcal{L}_{o} + \lambda_\text{det} \cdot \mathcal{L}_{det},
\end{aligned}
\end{equation}
where $\mathcal{L}_{h}$, $\mathcal{L}_{g}$, and $\mathcal{L}_{c}$ are losses of head heatmaps, gaze heatmaps, and connection maps. $\mathcal{L}_\text{det}$ is the loss of head detection maps, and $\mathcal{L}_{o}$ is the loss of out-of-frame flags. $\lambda_{h}$, $\lambda_{g}$, $\lambda_{c}$, $\lambda_{o}$, and $\lambda_\text{det}$ denotes the weights for each loss term.

For the head heatmap loss, we calculate the pixel-wise Mean Squared Error (MSE) between the predicted head heatmap $\{h_1, h_2, ..., h_M\} \in \textbf{H}$ and the ground-truth $\textbf{H}^\text{gt} = \{h^\text{gt}_1, h^\text{gt}_2, ..., h^\text{gt}_M\}$ as:

\begin{equation} 
\begin{aligned}
\mathcal{L}_{h}=\dfrac{1}{M}\dfrac{1}{mn}\sum_{k=1}^{M}\sum_{i=1}^{m}\sum_{j=1}^{n}[ h^\text{gt}_k(i,j) - h_k(i,j)]^2,
\end{aligned}
\end{equation}
where $M$ is the number of ground-truth head-target instances in the input image, $m$ and $n$ are heatmap width and height, and $h_k(i,j)$ is the pixel value of the k-th head heatmap at location $(i,j)$. The calculation of the MSE loss for $\mathcal{L}_{g}$ and $\mathcal{L}_{c}$ are the same as $\mathcal{L}_{h}$ with corresponding heatmaps.

We also employ MSE loss for the head detection map $\mathcal{L}_\text{det}$. Since there is just one head detection map per scene image, the calculation becomes:
\begin{equation} 
\begin{aligned}
\mathcal{L}_\text{det}=\dfrac{1}{mn}\sum_{i=1}^{m}\sum_{j=1}^{n}[ h_\text{det}^\text{gt}(i,j) - h_\text{det}(i,j)]^2.
\end{aligned}
\end{equation}

Lastly, the out-of-frame flag loss $\mathcal{L}_{o}$ is defined as the binary cross-entropy loss between the prediction $\{o_1, o_2, ..., o_M\} \in O$ and the ground-truth $O^\text{gt}=\{o^\text{gt}_1, o^\text{gt}_2, ..., o^\text{gt}_M\}$. The calculation is:
\begin{equation} 
\begin{aligned}
\mathcal{L}_{o} = - \dfrac{1}{M} \sum_{k=1}^{M}[ o^\text{gt}_k\log(o_k) + (1 - o_k^\text{gt}) \log(1 - o_k)].
\end{aligned}
\end{equation}

%% file: 05_experiment.tex
\section{Experiments}

\begin{table*}[t]
\centering
\small
\begin{tabularx}{\textwidth}{l *{8}{Y}} 
\toprule
\multirow{2}[2]{*}{Method} & \multicolumn{4}{c}{GazeFollow}  & \multicolumn{4}{c}{VideoAttentionTarget} \\ \cmidrule(lr){2-5} \cmidrule(lr){6-9}
                        &AUC$\uparrow$   & Avg. Dist.$\downarrow$ & Min. Dist.$\downarrow$ & mAP$\uparrow$ & AUC$\uparrow$  & Dist.$\downarrow$ & AP$\uparrow$ & mAP$\uparrow$ \\

\midrule
VideoAttention \cite{chong2020detecting}  & 0.921 & 0.137 & 0.077 & - & 0.860 & 0.134 & 0.853 & - \\
DAM \cite{fang2021dual}  & 0.922 & 0.124 & 0.067 & - & 0.905 & 0.108 & 0.896 & -\\
ESCNet \cite{bao2022escnet}  & 0.928 & 0.122 & - & - & 0.885 & 0.120 & 0.869 & - \\
Sharingan \cite{tafasca2023sharingan}  & \textbf{0.944} & 0.113 & 0.057 & - & - & 0.107 & 0.891 & - \\
\midrule
HGTTR \cite{tu2022end}   & 0.917  & 0.133 & 0.069 & 0.547 & 0.893  & 0.137  & 0.821 & 0.514 \\
Tonini et al. \cite{tonini2023object} & 0.922  & 0.072  & 0.033 & 0.573 & 0.923  & 0.102  & 0.944 & 0.607 \\
\midrule
\methodname  & 0.929 & \textbf{0.063} & \textbf{0.025} & \textbf{0.626} & \textbf{0.957}  & \textbf{0.064} & \textbf{1.000} & \textbf{0.772} \\
\bottomrule
\end{tabularx}
\caption{Comparison of \methodname to other SOTA gaze target detection methods on the GazeFollow and VideoAttentionTarget datasets in terms of gaze target heatmap AUC, gaze point distances (Avg. Dist., Min. Dist., Dist.), head-target instance mAP, and out-of-frame flag AP. \methodname outperforms both SOTA two-stage approaches (first to third rows), and SOTA multi-person end-to-end approaches (fourth to fifth rows) across all evaluation metrics.
}
\label{tab:compare_sota}
\end{table*}

\subsection{Datasets}
\textbf{GazeFollow} \cite{recasens2015they} is a large-scale dataset consisting of 122,143 images with 130,339 annotations on head-target instances. In total, there are 4,782 images with 44,191 head annotations. The gaze target of each head is annotated by ten individuals. We sample 2697 samples from the training set as the validation set. GazeFollow contains human faces and is collected from several public datasets, such as COCO \cite{lin2014microsoft} and Action 40 \cite{yao2011human}. There was no participant consent form since the data are from public data.

\noindent\textbf{VideoAttentionTarget} \cite{chong2020detecting} is a video-based gaze target dataset. It comprises 71,666 frames from 1,331 clips, with 164,514 annotations on head-target instances. Following the training procedure outlined in \cite{tu2022end,tonini2022multimodal,tonini2023object}, we adopt a sampling strategy on the training data, where one frame is selected from every five consecutive frames to avoid overfitting. We sampled 6 clips from the training set for validation purposes. 
The test set of this dataset consists of 2,771 frames from 10 clips with 6,377 annotations on head-target instances that remain unchanged in our experiments.
The data contain human faces and were retrieved from YouTube videos covering various sources including live interviews, sitcoms, reality shows, and movie clips. There was no participant consent form since the data are from public data.

\subsection{Ground-truth Generation.}
The ground-truth annotations from the GazeFollow and VideoAttentionTarget datasets only include head bounding boxes and the 2D positions of their corresponding gaze targets. 
Similarly, the pseudo labels obtained from RetinaFace \cite{deng2020retinaface} only include head bounding boxes.
To generate the ground-truth head heatmaps $\textbf{H}^\text{gt}$, gaze heatmaps $\textbf{G}^\text{gt}$, and head detection map $H_\text{det}^\text{gt}$, we assign values with Gaussian distributions centered at the head bounding box centers and gaze points.
For the ground-truth connection maps $\textbf{C}^\text{gt}$, we sample 50 points along the straight line connecting the head-target instance and then generate Gaussian distributions for each point. These Gaussian distributions collectively form connection maps $\textbf{C}^\text{gt}$.

\subsection{Implementation Details}
We use the pre-trained image encoder and denoising U-Net from Stable Diffusion \cite{rombach2022high} as the backbone. During training, the denoising U-Net is finetuned, while the image encoder is frozen without updating. We utilize AdamW optimizer \cite{loshchilov2018decoupled} along with a one-cycle-policy learning rate scheduler \cite{smith2019super}. For the GazeFollow dataset, the maximum learning rate is set to 3.2e-5 with 20 epochs of training. For the VideoAttentionTarget dataset, the maximum learning rate is set to 4e-4 with 10 epochs of training.
The weights for training loss $\lambda_{h}$, $\lambda_{g}$, $\lambda_{c}$, $\lambda_{o}$, $\lambda_\text{det}$ are set as 1.0, 2.5, 1.0, 1.0, and 1.0, respectively. The number of head-target proposals $N$ is set to 20 following previous works \cite{tonini2022multimodal,tu2022end}. The output heatmap width $m$ and height $n$ are both set to $64$.

\subsection{Evaluation Metrics}
Following previous works \cite{chong2020detecting,recasens2015they,tonini2023object,tu2022end}, we use the \textit{Area Under Curve} (\textbf{AUC}) and \textit{gaze point distance} (\textbf{Dist.}) metrics for evaluation. These two metrics are assessed exclusively when the ground-truth gaze targets are labeled as ``in-frame''.
AUC indicates the confidence level of the predicted gaze target locations with respect to the ground-truths. Gaze point distance measures the Euclidean distance between the predicted gaze points and the ground-truths, and it is normalized by the image size. The predicted gaze points are inferred from the location with the maximum value in the gaze heatmaps $\textbf{G}$. 

We employ \textit{mean Average Precision} (\textbf{mAP}) to assess the predicted head-target instances, following the setting in \cite{tu2022end}. A head-target instance is regarded as true positive if and only if the Intersection-Over-Union (IOU) ratio between the predicted and ground-truth head boxes is greater than 0.5, and the normalized gaze distance is less than 0.15. We extract the head bounding box with heatmap thresholding \cite{otsu1979threshold} followed by contour finding on $\textbf{H}$. Only the head IOU is considered if gaze targets are labeled as out-of-frame.

For the GazeFollow dataset, both the \textit{average distance} (\textbf{Avg. Dist.}) and \textit{minimum distance} (\textbf{Min. Dist.}) are reported due to multiple annotation instances for each given head. 
As for the out-of-frame gaze target instances in the VideoAttentionTarget dataset, we report \textit{Average Precision} (\textbf{AP}) on the out-of-frame flags in our evaluation. This metric does not apply to the GazeFollow dataset since the test set only includes in-frame gaze targets. 

\subsection{Comparison with State-of-the-Art}

The quantitative comparison between previous gaze target detection approaches and \methodname on the GazeFollow and VideoAttentionTarget datasets is summarized in \cref{tab:compare_sota}. It can be seen from the table that \methodname outperforms prior works across most of the evaluation metrics with large margins.
In contrast to conventional approaches where the understanding of heads and gaze targets is typically acquired independently before integration \cite{chong2020detecting,fang2021dual,bao2022escnet,tafasca2023sharingan}, \methodname provides a more unified solution. The advancements stress the importance of concurrently learning the association between heads and gaze targets. 
Note it is not a strictly fair comparison between our method and these two-stage approaches since these methods use the ground-truth head bounding boxes as input and there is no bipartite matching in the evaluation.
Compared to end-to-end approaches \cite{tu2022end,tonini2023object}, \methodname particularly demonstrates significant improvements on mAP, which is specifically designed for measuring head-target association. This shows the effectiveness in accurately identifying both individuals and their associated gaze targets, further highlighting the superiority of \methodname in modeling the nuanced correlation between heads and gaze targets.

It is noteworthy that the VideoAttentionTarget dataset is known for its complexity due to the high number of images containing multiple head-target annotations. This characteristic presents a challenge for accurately modeling interactions between multiple individuals and gaze targets. In contrast, the test set of the GazeFollow dataset only contains one in-frame head-target instance per image.
Compared to the current end-to-end SOTA \cite{tonini2023object}, the improvements achieved by \methodname on the VideoAttentionTarget dataset (3\% in AUC, 37\% in Dist., and 27\% in mAP) are relatively higher than those on the GazeFollow dataset (0.75\% in AUC, 12\% in Avg. Dist., and 9\% in mAP).
It explicitly demonstrates the efficiency of the integration of the proposed visual association and head feature re-injection techniques.

\subsection{Ablation Study}
\begin{table*}[t]
\centering
\small
\begin{tabularx}{\textwidth}{*{11}{Y}}
\toprule
\multirow{2}[2]{*}{Backbone} & \multirow{2}[2]{*}{$C$} & \multirow{2}[2]{*}{$F_\text{head}$}  &\multicolumn{4}{c}{GazeFollow}  & \multicolumn{4}{c}{VideoAttentionTarget} \\ \cmidrule(lr){4-7} \cmidrule(lr){8-11}
              &  &        & AUC$\uparrow$  & Avg. Dist.$\downarrow$ & Min. Dist.$\downarrow$  & mAP$\uparrow$ & AUC$\uparrow$ & Dist.$\downarrow$ & AP$\uparrow$  & mAP$\uparrow$  \\ 
\midrule
             \multirow{4}{*}{\begin{tabular}{@{}c@{}}Stable \\ Diffusion \cite{rombach2022high}\end{tabular}
             } & \xmark & \xmark & 0.910 & 0.084 & 0.042 & 0.500 & 0.951 & 0.079 & \textbf{1.000} & 0.759\\
              & \cmark & \xmark & \textbf{0.931} & 0.064 & 0.027 & 0.601 & 0.956 & 0.066 & \textbf{1.000} & 0.753 \\
               
             & \xmark & \cmark & 0.928 & 0.068 & 0.029 & 0.567 & 0.937 & 0.092 & \textbf{1.000} & 0.769   \\

              & \cmark & \cmark & 0.929 & \textbf{0.063} & \textbf{0.025} & \textbf{0.626} & \textbf{0.957}  & \textbf{0.064} & \textbf{1.000} & \textbf{0.772}  \\
\midrule
 \multirow{1}{*}{\begin{tabular}{@{}c@{}} DETR \cite{carion2020end}\end{tabular}} & \cmark & \cmark & 0.928 & 0.064 & 0.028 & 0.607 & 0.938 & 0.080  & \textbf{1.000} & 0.678  \\
\bottomrule
\end{tabularx}
\caption{Ablation study on model components and backbones. We evaluate the head feature re-injection (denoted as $F_\text{head}$) and connection maps (denoted as $C$) to show their effectiveness on two datasets. We switch the backbone from Stable Diffusion to DETR to demonstrate that the model components are not restricted to the diffusion model backbone.}
\label{tab:ablation_components}
\end{table*}

\subsubsection{Model components.}

We further investigate the incremental effectiveness of individual model components within \methodname, namely the head feature re-injection and the connection map. The results of this ablation study are presented in the first four rows of \cref{tab:ablation_components}, with the forth row showing the performance of our complete model \methodname. Note that the baseline model without the head feature re-injection nor the connection map is in the first row, which is essentially the VPD \cite{zhao2023unleashing} without the text prompt.

We consistently observe improvements in both average gaze point distances and mAP over the baseline when incorporating both the connection map and head feature re-injection. On the GazeFollow dataset, we observe improvements of 25\% (0.063 vs. 0.084) in Avg. Dist. and 24\% (0.626 vs. 0.500) in mAP. 
On the VideoAttentionTarget dataset, 18\% (0.064 vs. 0.079) in Dist. is achieved.
In general, we can see that employing either the connection map or the head feature re-injection can boost the performances across metrics except few exceptions. 
Overall, integrating both head feature re-injection and the connection map in \methodname yields superior performance 
compared to its variants.

\subsubsection{Backbones.}
Although \methodname leverages the semantic features extracted from the diffusion model, we expect that integrating the proposed network architecture with alternative backbone models can also yield improvements. To explore the generalizability of our model components, we substitute the backbone with a transformer-based object detector, DETR \cite{carion2020end}, as also employed in the previous gaze target detection SOTAs \cite{tu2022end,tonini2023object}. Note the remaining parts of the model including the head feature re-injection and the connection map are unchanged. 

The results of this experiment are presented in the last row of \cref{tab:ablation_components}.
In comparison to the state-of-the-art methods presented in \cref{tab:compare_sota}, our approach with the DETR backbone consistently outperforms across all evaluation metrics. More importantly, since we use the same backbone as in \cite{tu2022end,tonini2023object}, the improvements in performance solely come from the head feature re-injection and the connection map. This demonstrates the ability of the proposed components to generalize beyond a specific backbone.

Compared to \methodname that employs Stable Diffusion as the backbone (forth row in \cref{tab:compare_sota}), using DETR as the backbone achieves inferior performance.
It may be attributed to the fact that DETR is trained with a limited number of object classes, whereas Stable Diffusion is trained to generate realistic images, necessitating the learned features to encompass an understanding of diverse objects present in the training data.

\subsection{Qualitative Results}
\begin{figure}[t]
  \centering
  \includegraphics[width=\linewidth]{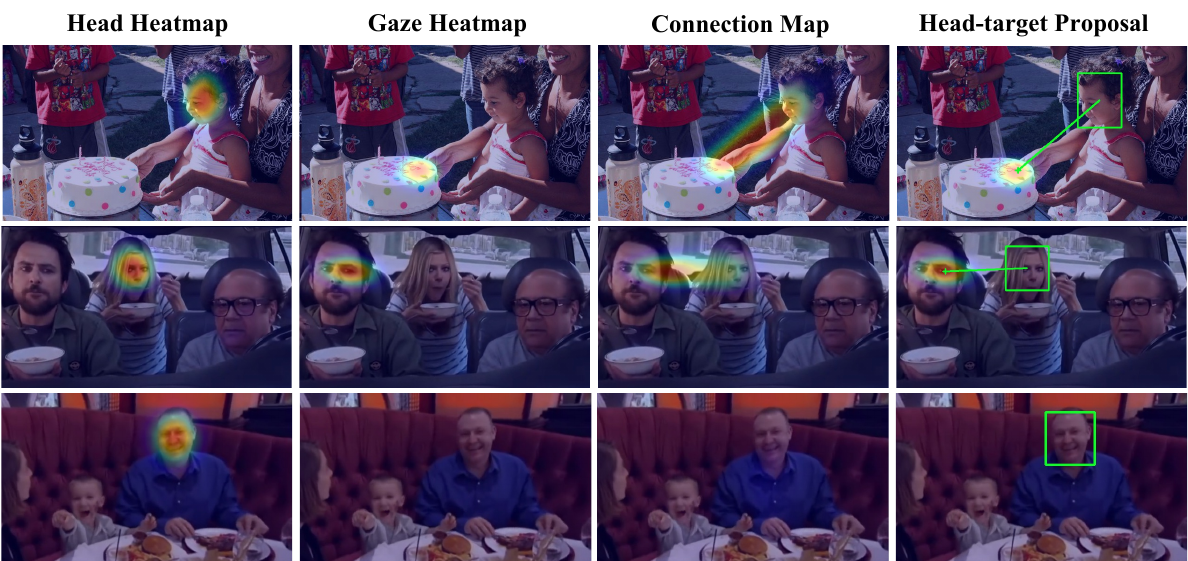}
  \caption{Predicted head heatmap, gaze heatmap, and the connection map from \methodname. 
  The first two rows demonstrate the strong associations between heads and in-frame gaze targets established by the connection maps. 
  The third row shows the comprehensive understanding of out-of-frame gaze targets in \methodname.
  }
  \label{fig:vis_ht}
\end{figure}

In \cref{fig:vis_ht}, we provide visualizations for the predicted head-target proposals, featuring their head heatmaps, gaze heatmaps, and connection maps. It is evident that the predicted connection maps establish strong associations between the heads and gaze targets. Notably, in the second row, the connection map effectively links the head and gaze target, even when the person is not facing directly towards the gaze target. In the third row, when the gaze target falls outside the frame, both the gaze heatmap and the connection map show low responses. This observation suggests the comprehensive understanding of gaze targets by \methodname in both in-frame and out-of-frame scenarios.

%% file: 07_conclusion.tex
\section{Discussion}
\noindent\textbf{Conclusion.}
We present \methodname, an end-to-end multi-person gaze target detection framework leveraging semantic features from a pre-trained diffusion model, improving head priors through head feature re-injection, and establishing explicit associations between heads and gaze targets with a connection map. \methodname achieves state-of-the-art performance in gaze target detection on two standard datasets.

\noindent\textbf{Limitation and Future Work.}
Although our experiments have shown the effectiveness of the proposed heatmap-based methodology, we believe that further improvements in head-target modeling could be achieved by formulating the problem to predict a single joint heatmap encompassing information for both heads and gaze targets. 
Similar to previous approaches, \methodname is also constrained by a predefined cap on the number $N$ of head-target instances per image.
Future work could explore the feasibility of allowing an arbitrary number of gaze-target instances to better align with real-world scenarios.

%% file: 00_main.bbl
\begin{thebibliography}{10}
\providecommand{\url}[1]{#1}
\csname url@rmstyle\endcsname
\providecommand{\newblock}{\relax}
\providecommand{\bibinfo}[2]{#2}
\providecommand\BIBentrySTDinterwordspacing{\spaceskip=0pt\relax}
\providecommand\BIBentryALTinterwordstretchfactor{4}
\providecommand\BIBentryALTinterwordspacing{\spaceskip=\fontdimen2\font plus
\BIBentryALTinterwordstretchfactor\fontdimen3\font minus \fontdimen4\font\relax}
\providecommand\BIBforeignlanguage[2]{{%
\expandafter\ifx\csname l@#1\endcsname\relax
\typeout{** WARNING: IEEEtran.bst: No hyphenation pattern has been}%
\typeout{** loaded for the language `#1'. Using the pattern for}%
\typeout{** the default language instead.}%
\else
\language=\csname l@#1\endcsname
\fi
#2}}

\bibitem{admoni2017social}
H.~Admoni and B.~Scassellati, ``Social eye gaze in human-robot interaction: a review,'' \emph{Journal of Human-Robot Interaction}, vol.~6, no.~1, pp. 25--63, 2017.

\bibitem{schellen2021robot}
E.~Schellen, F.~Bossi, and A.~Wykowska, ``Robot gaze behavior affects honesty in human-robot interaction,'' \emph{Frontiers in Artificial Intelligence}, vol.~4, p. 663190, 2021.

\bibitem{sheikhi2015combining}
S.~Sheikhi and J.-M. Odobez, ``Combining dynamic head pose--gaze mapping with the robot conversational state for attention recognition in human--robot interactions,'' \emph{Pattern Recognition Letters}, vol.~66, pp. 81--90, 2015.

\bibitem{arreghini2024predicting}
S.~Arreghini, G.~Abbate, A.~Giusti, and A.~Paolillo, ``Predicting the intention to interact with a service robot: the role of gaze cues,'' in \emph{2017 IEEE International Conference on Robotics and Automation (ICRA)}.\hskip 1em plus 0.5em minus 0.4em\relax IEEE, 2024.

\bibitem{oishi20214d}
S.~Oishi, K.~Koide, M.~Yokozuka, and A.~Banno, ``4d attention: Comprehensive framework for spatio-temporal gaze mapping,'' \emph{IEEE Robotics and Automation Letters}, vol.~6, no.~4, pp. 7240--7247, 2021.

\bibitem{jayasuriya20243d}
M.~Jayasuriya, G.~Hu, D.~D.~K. Le, K.~Ang, S.~Sankaran, and D.~Liu, ``A 3d vector field and gaze data fusion framework for hand motion intention prediction in human-robot collaboration,'' in \emph{2024 IEEE International Conference on Robotics and Automation (ICRA)}.\hskip 1em plus 0.5em minus 0.4em\relax IEEE, 2024, pp. 5637--5643.

\bibitem{zhang2023human}
Q.~Zhang, Z.~Hu, Y.~Song, J.~Pei, and J.~Liu, ``The human gaze helps robots run bravely and efficiently in crowds,'' in \emph{2023 IEEE International Conference on Robotics and Automation (ICRA)}.\hskip 1em plus 0.5em minus 0.4em\relax IEEE, 2023, pp. 7540--7546.

\bibitem{holman2021watch}
B.~Holman, A.~Anwar, A.~Singh, M.~Tec, J.~Hart, and P.~Stone, ``Watch where you’re going! gaze and head orientation as predictors for social robot navigation,'' in \emph{2021 IEEE International Conference on Robotics and Automation (ICRA)}.\hskip 1em plus 0.5em minus 0.4em\relax IEEE, 2021, pp. 3553--3559.

\bibitem{prada2023gaze}
J.~D.~P. Prada, M.~H. Lee, and C.~Song, ``A gaze-speech system in mixed reality for human-robot interaction,'' in \emph{2023 IEEE International Conference on Robotics and Automation (ICRA)}.\hskip 1em plus 0.5em minus 0.4em\relax IEEE, 2023, pp. 7547--7553.

\bibitem{razali2022using}
H.~Razali and Y.~Demiris, ``Using eye gaze to forecast human pose in everyday pick and place actions,'' in \emph{2022 International Conference on Robotics and Automation (ICRA)}.\hskip 1em plus 0.5em minus 0.4em\relax IEEE, 2022, pp. 8497--8503.

\bibitem{qian2023gvgnet}
K.~Qian, Z.~Zhang, W.~Song, and J.~Liao, ``Gvgnet: Gaze-directed visual grounding for learning under-specified object referring intention,'' \emph{IEEE Robotics and Automation Letters}, 2023.

\bibitem{zhang2021foot}
K.~Zhang, H.~Liu, Z.~Fan, X.~Chen, Y.~Leng, C.~W. de~Silva, and C.~Fu, ``Foot placement prediction for assistive walking by fusing sequential 3d gaze and environmental context,'' \emph{IEEE Robotics and Automation Letters}, vol.~6, no.~2, pp. 2509--2516, 2021.

\bibitem{kim2020using}
H.~Kim, Y.~Ohmura, and Y.~Kuniyoshi, ``Using human gaze to improve robustness against irrelevant objects in robot manipulation tasks,'' \emph{IEEE Robotics and Automation Letters}, vol.~5, no.~3, pp. 4415--4422, 2020.

\bibitem{kshirsagar2020robot}
A.~Kshirsagar, M.~Lim, S.~Christian, and G.~Hoffman, ``Robot gaze behaviors in human-to-robot handovers,'' \emph{IEEE Robotics and Automation Letters}, vol.~5, no.~4, pp. 6552--6558, 2020.

\bibitem{zhang2017mpiigaze}
X.~Zhang, Y.~Sugano, M.~Fritz, and A.~Bulling, ``Mpiigaze: Real-world dataset and deep appearance-based gaze estimation,'' \emph{IEEE TPAMI}, vol.~41, no.~1, pp. 162--175, 2017.

\bibitem{yu2018deep}
Y.~Yu, G.~Liu, and J.-M. Odobez, ``Deep multitask gaze estimation with a constrained landmark-gaze model,'' in \emph{Proceedings of the European conference on computer vision (ECCV) workshops}, 2018, pp. 0--0.

\bibitem{zhou2017two}
X.~Zhou, H.~Cai, Y.~Li, and H.~Liu, ``Two-eye model-based gaze estimation from a kinect sensor,'' in \emph{2017 IEEE International Conference on Robotics and Automation (ICRA)}.\hskip 1em plus 0.5em minus 0.4em\relax IEEE, 2017, pp. 1646--1653.

\bibitem{cheng2023dvgaze}
Y.~Cheng and F.~Lu, ``Dvgaze: Dual-view gaze estimation,'' in \emph{ICCV}, 2023, pp. 20\,632--20\,641.

\bibitem{bandi2023new}
C.~Bandi and U.~Thomas, ``A new efficient eye gaze tracker for robotic applications,'' in \emph{2023 IEEE International Conference on Robotics and Automation (ICRA)}.\hskip 1em plus 0.5em minus 0.4em\relax IEEE, 2023, pp. 6153--6159.

\bibitem{kratzer2020mogaze}
P.~Kratzer, S.~Bihlmaier, N.~B. Midlagajni, R.~Prakash, M.~Toussaint, and J.~Mainprice, ``Mogaze: A dataset of full-body motions that includes workspace geometry and eye-gaze,'' \emph{IEEE Robotics and Automation Letters}, vol.~6, no.~2, pp. 367--373, 2020.

\bibitem{recasens2015they}
A.~Recasens, A.~Khosla, C.~Vondrick, and A.~Torralba, ``Where are they looking?'' \emph{Advances in neural information processing systems}, vol.~28, 2015.

\bibitem{recasens2017following}
A.~Recasens, C.~Vondrick, A.~Khosla, and A.~Torralba, ``Following gaze in video,'' in \emph{Proceedings of the IEEE International Conference on Computer Vision}, 2017, pp. 1435--1443.

\bibitem{chong2018connecting}
E.~Chong, N.~Ruiz, Y.~Wang, Y.~Zhang, A.~Rozga, and J.~M. Rehg, ``Connecting gaze, scene, and attention: Generalized attention estimation via joint modeling of gaze and scene saliency,'' in \emph{Proceedings of the European conference on computer vision (ECCV)}, 2018, pp. 383--398.

\bibitem{lian2018believe}
D.~Lian, Z.~Yu, and S.~Gao, ``Believe it or not, we know what you are looking at!'' in \emph{Asian Conference on Computer Vision}.\hskip 1em plus 0.5em minus 0.4em\relax Springer, 2018, pp. 35--50.

\bibitem{chong2020detecting}
E.~Chong, Y.~Wang, N.~Ruiz, and J.~M. Rehg, ``Detecting attended visual targets in video,'' in \emph{Proceedings of the IEEE/CVF conference on computer vision and pattern recognition}, 2020, pp. 5396--5406.

\bibitem{bao2022escnet}
J.~Bao, B.~Liu, and J.~Yu, ``Escnet: Gaze target detection with the understanding of 3d scenes,'' in \emph{Proceedings of the IEEE/CVF Conference on Computer Vision and Pattern Recognition}, 2022, pp. 14\,126--14\,135.

\bibitem{tu2022end}
D.~Tu, X.~Min, H.~Duan, G.~Guo, G.~Zhai, and W.~Shen, ``End-to-end human-gaze-target detection with transformers,'' in \emph{2022 IEEE/CVF Conference on Computer Vision and Pattern Recognition (CVPR)}.\hskip 1em plus 0.5em minus 0.4em\relax IEEE, 2022, pp. 2192--2200.

\bibitem{tonini2023object}
F.~Tonini, N.~Dall'Asen, C.~Beyan, and E.~Ricci, ``Object-aware gaze target detection,'' in \emph{Proceedings of the IEEE/CVF International Conference on Computer Vision}, 2023, pp. 21\,860--21\,869.

\bibitem{chen2023diffusiondet}
S.~Chen, P.~Sun, Y.~Song, and P.~Luo, ``Diffusiondet: Diffusion model for object detection,'' in \emph{Proceedings of the IEEE/CVF International Conference on Computer Vision}, 2023, pp. 19\,830--19\,843.

\bibitem{ji2023ddp}
Y.~Ji, Z.~Chen, E.~Xie, L.~Hong, X.~Liu, Z.~Liu, T.~Lu, Z.~Li, and P.~Luo, ``Ddp: Diffusion model for dense visual prediction,'' \emph{arXiv preprint arXiv:2303.17559}, 2023.

\bibitem{rombach2022high}
R.~Rombach, A.~Blattmann, D.~Lorenz, P.~Esser, and B.~Ommer, ``High-resolution image synthesis with latent diffusion models,'' in \emph{Proceedings of the IEEE/CVF conference on computer vision and pattern recognition}, 2022, pp. 10\,684--10\,695.

\bibitem{saxena2024surprising}
S.~Saxena, C.~Herrmann, J.~Hur, A.~Kar, M.~Norouzi, D.~Sun, and D.~J. Fleet, ``The surprising effectiveness of diffusion models for optical flow and monocular depth estimation,'' \emph{NeurIPS}, vol.~36, 2024.

\bibitem{zhao2023unleashing}
W.~Zhao, Y.~Rao, Z.~Liu, B.~Liu, J.~Zhou, and J.~Lu, ``Unleashing text-to-image diffusion models for visual perception,'' \emph{arXiv preprint arXiv:2303.02153}, 2023.

\bibitem{fang2021dual}
Y.~Fang, J.~Tang, W.~Shen, W.~Shen, X.~Gu, L.~Song, and G.~Zhai, ``Dual attention guided gaze target detection in the wild,'' in \emph{Proceedings of the IEEE/CVF conference on computer vision and pattern recognition}, 2021, pp. 11\,390--11\,399.

\bibitem{tafasca2023childplay}
S.~Tafasca, A.~Gupta, and J.-M. Odobez, ``Childplay: A new benchmark for understanding children's gaze behaviour,'' in \emph{ICCV}, 2023, pp. 20\,935--20\,946.

\bibitem{tafasca2023sharingan}
------, ``Sharingan: A transformer-based architecture for gaze following,'' \emph{arXiv preprint arXiv:2310.00816}, 2023.

\bibitem{tonini2022multimodal}
F.~Tonini, C.~Beyan, and E.~Ricci, ``Multimodal across domains gaze target detection,'' in \emph{Proceedings of the 2022 International Conference on Multimodal Interaction}, 2022, pp. 420--431.

\bibitem{wang2024transgop}
B.~Wang, C.~Guo, Y.~Jin, H.~Xia, and N.~Liu, ``Transgop: Transformer-based gaze object prediction,'' \emph{arXiv preprint arXiv:2402.13578}, 2024.

\bibitem{zhao2020learning}
H.~Zhao, M.~Lu, A.~Yao, Y.~Chen, and L.~Zhang, ``Learning to draw sight lines,'' \emph{International Journal of Computer Vision}, vol. 128, pp. 1076--1100, 2020.

\bibitem{sohl2015deep}
J.~Sohl-Dickstein, E.~Weiss, N.~Maheswaranathan, and S.~Ganguli, ``Deep unsupervised learning using nonequilibrium thermodynamics,'' in \emph{International conference on machine learning}.\hskip 1em plus 0.5em minus 0.4em\relax PMLR, 2015, pp. 2256--2265.

\bibitem{ramesh2021dalle}
A.~Ramesh, S.~Goyal, and S.~Chintala, ``Dall·e: Creating images from text using a conditional transformer,'' \emph{arXiv preprint arXiv:2102.12092}, 2021.

\bibitem{saharia2022photorealistic}
C.~Saharia, W.~Chan, S.~Saxena, L.~Li, J.~Whang, E.~L. Denton, K.~Ghasemipour, R.~Gontijo~Lopes, B.~Karagol~Ayan, T.~Salimans, \emph{et~al.}, ``Photorealistic text-to-image diffusion models with deep language understanding,'' \emph{Advances in Neural Information Processing Systems}, vol.~35, pp. 36\,479--36\,494, 2022.

\bibitem{ramesh2021zero}
A.~Ramesh, M.~Pavlov, G.~Goh, S.~Gray, C.~Voss, A.~Radford, M.~Chen, and I.~Sutskever, ``Zero-shot text-to-image generation,'' in \emph{International Conference on Machine Learning}.\hskip 1em plus 0.5em minus 0.4em\relax PMLR, 2021, pp. 8821--8831.

\bibitem{clark2024text}
K.~Clark and P.~Jaini, ``Text-to-image diffusion models are zero shot classifiers,'' \emph{NeurIPS}, vol.~36, 2024.

\bibitem{yang2024depth}
L.~Yang, B.~Kang, Z.~Huang, X.~Xu, J.~Feng, and H.~Zhao, ``Depth anything: Unleashing the power of large-scale unlabeled data,'' \emph{CVPR}, 2024.

\bibitem{pnvr2023ld}
K.~Pnvr, B.~Singh, P.~Ghosh, B.~Siddiquie, and D.~Jacobs, ``Ld-znet: A latent diffusion approach for text-based image segmentation,'' in \emph{ICCV}, 2023, pp. 4157--4168.

\bibitem{carion2020end}
N.~Carion, F.~Massa, G.~Synnaeve, N.~Usunier, A.~Kirillov, and S.~Zagoruyko, ``End-to-end object detection with transformers,'' in \emph{European conference on computer vision}.\hskip 1em plus 0.5em minus 0.4em\relax Springer, 2020, pp. 213--229.

\bibitem{kuhn1955hungarian}
H.~W. Kuhn, ``The hungarian method for the assignment problem,'' \emph{Naval research logistics quarterly}, vol.~2, no. 1-2, pp. 83--97, 1955.

\bibitem{lin2014microsoft}
T.-Y. Lin, M.~Maire, S.~Belongie, J.~Hays, P.~Perona, D.~Ramanan, P.~Doll{\'a}r, and C.~L. Zitnick, ``Microsoft coco: Common objects in context,'' in \emph{Computer Vision--ECCV 2014: 13th European Conference, Zurich, Switzerland, September 6-12, 2014, Proceedings, Part V 13}.\hskip 1em plus 0.5em minus 0.4em\relax Springer, 2014, pp. 740--755.

\bibitem{yao2011human}
B.~Yao, X.~Jiang, A.~Khosla, A.~L. Lin, L.~Guibas, and L.~Fei-Fei, ``Human action recognition by learning bases of action attributes and parts,'' in \emph{2011 International conference on computer vision}.\hskip 1em plus 0.5em minus 0.4em\relax IEEE, 2011, pp. 1331--1338.

\bibitem{deng2020retinaface}
J.~Deng, J.~Guo, E.~Ververas, I.~Kotsia, and S.~Zafeiriou, ``Retinaface: Single-shot multi-level face localisation in the wild,'' in \emph{Proceedings of the IEEE/CVF conference on computer vision and pattern recognition}, 2020, pp. 5203--5212.

\bibitem{loshchilov2018decoupled}
I.~Loshchilov and F.~Hutter, ``Decoupled weight decay regularization,'' in \emph{International Conference on Learning Representations}, 2018.

\bibitem{smith2019super}
L.~N. Smith and N.~Topin, ``Super-convergence: Very fast training of neural networks using large learning rates,'' in \emph{Artificial intelligence and machine learning for multi-domain operations applications}, vol. 11006.\hskip 1em plus 0.5em minus 0.4em\relax SPIE, 2019, pp. 369--386.

\bibitem{otsu1979threshold}
N.~Otsu, ``A threshold selection method from gray-level histograms,'' \emph{IEEE transactions on systems, man, and cybernetics}, vol.~9, no.~1, pp. 62--66, 1979.

\end{thebibliography}
